\documentclass{article}
\usepackage[
backend=biber,
style=numeric,
sorting=ynt
]{biblatex}
\usepackage{amssymb}
\usepackage{amsmath}
\usepackage{bm}
\usepackage{color}
\DeclareMathOperator{\Tr}{Tr}
\usepackage{hyperref}

\usepackage{geometry}
 \usepackage[acronym]{glossaries}

\usepackage{optidef}
\makeglossaries

\newglossaryentry{entryOne}
{
        name=Glossary Entry,
        description={Glossary entries are used to provide definitions for words in your document}
}

\title{An introduction to optimization under uncertainty -- A short survey}
\author{Keivan Shariatmadar$^2$ Kaizheng Wang$^2$  Calvin R. Hubbard$^1$ Hans Hallez$^2$  David Moens$^1$\\
{\small $^1$ Department of Mechanical Engineering, LMSD, Campus De Nayer, KU Leuven, Belgium}\\
{\small $^2$ Department of Computer Science, M-Group, Campus Bruges, KU Leuven, Belgium }
}
\date{\small\it\{keivan.shariatmadar,kaizheng.wang,rory.hubbard,hans.hallez,david.moens\}@kuleuven.be}

\begin{document}

\maketitle

\begin{abstract}

Optimization equips engineers and scientists in a variety of fields with the ability to transcribe their problems into a generic formulation and receive optimal solutions with relative ease.
Industries ranging from aerospace to robotics continue to benefit from advancements in optimization theory and the associated algorithmic developments.
Nowadays, optimization is used in real time on autonomous systems acting in safety critical situations, such as self-driving vehicles.
It has become increasingly more important to produce robust solutions by incorporating uncertainty into optimization programs.
This paper provides a short survey about the state of the art in optimization under uncertainty.
The paper begins with a brief overview of the main classes of optimization without uncertainty.
The rest of the paper focuses on the different methods for handling both aleatoric and epistemic uncertainty.
Many of the applications discussed in this paper are within the domain of control.
The goal of this survey paper is to briefly touch upon the state of the art in a variety of different methods and refer the reader to other literature for more in-depth treatments of the topics discussed here.

\end{abstract} \hspace{10pt}

\section{Introduction}

Optimization is a vast field and is arguably one of the most useful tools for scientists and engineers.
With applications in almost any industry, from operations research to climate analysis to process control to robotics, the need to further our understanding of optimization and develop efficient algorithms to solve optimization problems is clear.
The mathematical structure and geometric interpretations of optimization make it an exciting academic research area.
It is interesting for its own sake.
So it is fortunate that optimization also happens to be extremely useful in solving real problems and developing real technology.
Another fortunate feature of optimization is that it has a rich history of remarkable leaps in understanding.
One discovery of particular importance was the realization that the distinction between complex and easy optimization problems does not hinge on linearity, but rather, convexity \cite{optimality-rockafellar}.
Rockafellar published this historical paper in 1993.
The date of his seminal discovery is interesting to note when put into context.
Humans first stepped on the moon in 1969.
So it wasn't until over 20 years later that we realized the fundamental importance of convexity in optimization problems.
Nowadays, we are consistently sending rockets to space and back, which would not be possible without numerical optimization, in particular, convex optimization \cite{rocket-landing}.
How many more discoveries of the same magnitude as Rockafellar's are left to make?
Currently, it seems that the theory behind convex optimization is nearly complete.
So what developments are necessary to further our understanding of optimization and increase its utility?
Optimization that includes uncertainty is a research frontier that is ripe for research.

In this survey paper, we review deterministic optimization and optimization under both aleatoric and epistemic uncertainty. From our past research about modeling uncertainty and optimization under uncertainty, we start applying them to the Artificial Intelligence (AI) domain. We have realized that optimization under uncertainty is one of the important filed in AI. We start to first do a literature study and explain what (important) methods have been used in optimization through this survey paper.

The structure of this paper is as follows:
In Section 2 we briefly review optimization without uncertainty, convex and nonconvex.
In Section 3, we review the state-of-the-art methods for optimization under aleatoric and epistemic uncertainty.
In Section 4, we discuss optimization under uncertainty broadly and compare the different approaches.
In Section 5, we conclude with a brief summary and possible research directions.
Throughout the paper, we provide specific applications of optimization where many of the applications are focused on optimal control.
It is important to remember, however, that these applications are just one of many use cases for the techniques discussed in this survey paper.
We provide these applications for concreteness.
For a complete survey of optimization as applied to optimal control specifically, we refer the reader to the excellent review paper \cite{trajectory-generation}.

\section{Deterministic Optimization}
\textcolor{black}{In this section, we review the main two classes of deterministic Optimization, namely convex and non-convex optimization, respectively.}
\subsection{Convex Optimization}

A generic optimization problem can be written as:

\begin{mini}|s|[0]
    {x \in X}{J(x)}
    {} 
    {\label{eq:minimizationProblem}}{}
    \addConstraint{g(x)}{ \le 0}
    \addConstraint{h(x)}{ = 0}
\end{mini}

where $x$ is the decision variable vector, $J$ is the objective function, $g$ are inequality constraints, and $h$ are equality constraints.
If $g$ and $h$ are removed, the problem becomes an unconstrained optimization problem.
The goal in constrained optimization is to minimize $J$ while satisfying the constraints imposed by $g$ and $h$. \textcolor{black}{A wide variety of problems in the real world can be transcribed into the above form.}

In convex optimization, $J$ is a convex function, $g$ creates a convex set, and $h$ is an affine function.
For formal definitions of convex functions and convex sets we refer the reader to the classical book, \cite{convex-optimization-book}.
The key feature of convex optimization problems is that global conclusions can be made from local function evaluations.
This property is what makes convex optimization an important area of study and is why it can be effectively applied to real problems.
This allows solvers to quickly and efficiently find the true, globally optimal solution to the problem up to arbitrary precision (discounting floating point precision limits imposed by computers).

There are four main classes of convex optimization problems: linear programs, quadratic programs, second-order cone programs, and semidefinite programs \cite{convex-optimization-book}.
In linear programs, a linear objective is optimized over a polyhedron, which is the shape of the feasible set of linear programs.
Written in standard form, the feasible space of a linear program is the intersection of an affine subspace and the nonnegative orthant.

\begin{mini}|s|[0]
    {x \in X}{c'x + d}
    {} 
    {\label{eq:minimizationProblem}}{}
    \addConstraint{x}{ \ge 0}
    \addConstraint{Ax}{ = b}
\end{mini}

A nice property of linear programs is that an optimal solution (if one exists) can always be found at one of the vertices of the polyhedron.
This enables solution methods such as the simplex method to give algebraic solutions which can then be used for sensitivity analysis.

Quadratic programs have the same feasible region description as linear programs. The difference is the objective takes a quadratic form.

\begin{mini}|s|[0]
    {x \in X}{(1/2)x'Qx + c'x + d}
    {} 
    {\label{eq:minimizationProblem}}{}
    \addConstraint{Gx}{ \le h}
    \addConstraint{Ax}{ = b}
\end{mini}

For this reason, optimal solutions to quadratic programs are not always found on a vertex of the polyhedron feasible region. 
Quadratic programs are often used for model predictive controllers where the constraints define the system dynamics and control limits and the objective specifies a cost that penalizes state error and control effort.

Second-order cone programs optimize a linear objective over a feasible region that is specified by the intersection of an affine subspace and the second-order cone, also referred to as the Lorentz cone \cite{Lorentz-cone}.

\begin{mini}|s|[0]
    {x \in X}{q'x}
    {} 
    {\label{eq:minimizationProblem}}{}
    \addConstraint{|| Gx + h ||_2}{ \le c'x + d}
    \addConstraint{Ax}{ = b}
\end{mini}

Robust linear programs can be cast as second-order cone programs 
\cite{second-order-cone-programming-applications}.

Semidefinite programs optimize a linear objective over the intersection of an affine subspace and the cone of symmetric positive semidefinite matrices.
The feasible space of a semidefinite program is a spectrahedra \cite{semidefinite-optimization-convex-algebraic-geometry}.

\begin{mini}|s|[0]
    {X \in S^n}{\langle C, x \rangle}
    {} 
    {\label{eq:minimizationProblem}}{}
    \addConstraint{\langle A, X \rangle}{= b}
    \addConstraint{X}{ \succeq 0}
\end{mini}

where $\langle X, Y \rangle := \Tr(X^TY)$.
A unifying trait of the four convex optimization problems discussed above is that the feasible region lies within a proper cone.
Problems that optimize over the intersection of an affine subspace and a convex cone are referred to as conic programs, which subsume the four types of optimization problems mentioned above.

\subsection{Nonconvex Optimization}
Many of the real-world problems that we care about have nonconvexities either in the objective or in the constraints.
We briefly discuss two different approaches to solving nonconvex optimization problems in this section.

Lossless convex relaxations: It is sometimes possible to remove acute nonconvexities in a lossless manner, meaning that the problem can be reconfigured to be a convex program where the optimal solution to the convexified problem is the same as the optimal solution to the original problem.

\textcolor{black}{\cite{convex-approach-mars-powered-descent} shows that the nonconvex thrust vector $T(t) \in \mathbb{R}^3$ constraint of a powered descent space vehicle (described in \eqref{18}) can be removed in a lossless way by introducing additional decision variables, given in \eqref{19}.}
\begin{equation} \label{18}
    \rho_{min} \leq || T(t) ||_2 \leq \rho_{max}, \forall t \in [0, t_f]
\end{equation}
\begin{equation} \label{19}
    \rho_{min} \leq \sigma(t) \leq \rho_{max}, || T(t) ||_2 \leq \sigma(t), \forall t \in [0, t_f]
\end{equation}
It is proofed via the maximum principle that the optimal solution to the lifted problem \eqref{19} can always be projected down to the feasible region defined by the original coordinates in \eqref{18}.

Sequential convex programming: Methods for solving nonconvex problems to local optimality have been widely studied.
The general approach is to iteratively approximate the original problem with linearizations and other convex relaxations, solve the approximate convex problem, project the solution from the convex subproblem back to the feasible space of the nonconvex problem, and repeat the process starting from the new projected point until some convergence criteria.
See \cite{convex-optimization-for-trajectory-generation} for a detailed review of a couple of specific algorithms for sequential convex programming.

Also worth mentioning are mixed-integer programs which are problems with decision variables that are restricted to be integer-valued.
These problems are nonconvex due to the integer variables but if the rest of the problem is specified by a convex objective and convex constraints, then these problems can still be efficiently solved to global optimality via branch-and-bound algorithms \cite{branch-and-bound}.
Mixed-integer programs have the ability to combine discrete, combinatorial aspects of problems with smooth constraints.
Mixed-integer programs are especially interesting when used within a learning framework \cite{lvis-contact-aware-controllers}.

\section{Optimization under uncertainty}

\textbf{Aleatoric} uncertainty is the uncertainty resulting from true randomness in a given process.
This uncertainty cannot be reduced by further experimentation.
Aleatoric uncertainty is typically modeled by probability distributions.
One of the most common models for aleatoric uncertainty is the gaussian probability distribution.

\textbf{Epistemic} uncertainty stems from a lack of knowledge about the system or process of interest.
This type of uncertainty can be reduced by obtaining further information.
One canonical example of a model for epistemic uncertainty is the probability box or p-box \cite{shariatmadar2019pbox}.
The p-box is defined by lower and upper cumulative distribution functions.
The true distribution function is located within these lower and upper bounds.

In this section, we review the main classes of optimization methods that handle uncertainty.

\subsection{Robust optimization}

Robust optimization optimizes for the worse case.
It does not require any specified probability distributions of the uncertain data.
Instead, in robust optimization, one bounds the uncertain variables to a set of possible values and then optimizes for the worst possible realization of the uncertain variables from those sets.
A generic formulation of the robust optimization problem can be written as follows:

\begin{mini}|s|[0]
    {x} {f(x)}
    {\label{0}}{}
    \addConstraint{g(x, \Delta)}{\leq 0}{, \; \forall \Delta \in \bm{\Delta}}
    \addConstraint{h(x)}{= 0}
\end{mini}

where $\bm{\Delta}$ is the set of all possible uncertainties in the problem.
The structure of $\bm{\Delta}$ significantly impacts the solution approach and overall tractability of the problem.
See \cite{optimization-uncertainty-survey} for a condensed list of common uncertainty sets used in robust optimization problems.

In the context of optimal control, the uncertainty typically comes from uncertainty in the parameters of the system dynamics.
A simple model of uncertainty in dynamic systems is bounded external additive disturbance:

\begin{equation} \label{uncertain_linear_system}
    \dot{x}(t) = A(t)x(t) + B(t)u(t) + Gw(t), \; w(t) \in \bm{W}
\end{equation}

where the disturbance set $\bm{W}$ is usually a convex, compact set.

Using this model, the robust optimal control problem can be solved via a minimax formulation:

\begin{equation*}
\begin{aligned}
& \underset{x, u}{\text{min}} \; \underset{w}{\text{max}}
& & \int_{0}^{T} l(x(t), u(t)) \; dt  + V_f(x(T))\\
& \text{subject to}
& & u(t) \in U, \; \forall w \in \bm{W}\\
& & & x(t) \in X, \; \forall w \in \bm{W}\\
\end{aligned}
\end{equation*}

Another popular approach is to assume the state space matrices $A$ and $B$ come from polytopic sets:

\begin{equation} \label{uncertain_linear_system}
    \dot{x}(t) = A(t)x(t) + B(t)u(t)
\end{equation}

\begin{equation} \label{uncertain_linear_system}
    A(t) \in Co(A_1, ..., A_n), \; B(t) \in Co(B_1, ..., B_n)
\end{equation}

where $Co$ denotes the convex hull.

See \cite{robust-design-survey} for an in-depth review of robust optimization.

\subsubsection{Sum of squares (SOS) optimization}
Sum of squares optimization is an active research area with applications in machine learning, control theory, and several other disciplines.
It can be seen as a particular type of robust optimization when applied to systems analysis and control.
An important problem in mathematics is checking the global nonnegativity of a function of multiple variables:

\begin{equation} \label{1}
    F(x) \geq 0 \ \forall x
\end{equation}

In the general case, the problem can be shown to be undecidable.
To make the problem tractable yet still useful, it is constructive to consider the class of polynomial functions and for polynomial functions, a sufficient condition to show global nonnegativity is to construct a sum of squares decomposition of the polynomial:

\begin{equation} \label{2}
    F(x) = \sum_{i}f_i^2(x)
\end{equation}

A polynomial that is the sum of squares can also be expressed in the quadratic form:

\begin{equation} \label{3}
    F(x) = z'Qz
\end{equation}

where $Q$ is positive semi-definite and $z$ is a basis of monomials of degree less than or equal to half of the degree of $F$ \cite{parrilo_thesis}.
After selecting a monomial basis vector, searching for a positive semi-definite $Q$ can be done via semidefinite programming for which known efficient algorithms exist \cite{boyd-semidefinite-programming}.

One example of how the sum of squares optimization is used for optimization under uncertainty is in the analysis and synthesis of Lyapunov stable systems with bounded uncertainty in either the system dynamics or operating environment.
The authors in \cite{funnel-libraries} utilize the sum of squares programming to generate trajectories for a fixed-wing aircraft that are guaranteed to succeed while taking into account the possibility of bounded disturbances, uncertainty in the environment, and uncertainty in the parametric model.
Their approach focuses on computing tight approximations of the reachable sets that the system may evolve to over the course of a trajectory.
So for a closed loop time-varying system defined in error coordinates around a trajectory and using uncertainty $w(t)$ from a semi-algebraic set $\{w | g_{w,j}(w) \geq 0, \forall j = 1,...,N_w\}$ to model external disturbances or parametric model uncertainties (aleatoric and epistemic uncertainty):

\begin{equation} \label{4}
    \dot{x} = f(t, x(t), w(t))
\end{equation}

they parameterize the reachable sets $F(t)$ by the sublevel sets of nonnegative time-varying functions $V$:

\begin{equation} \label{5}
    F(t) = \{x(t) | V(t, x) \leq \rho(t)\}
\end{equation}

and thus the constraint:

\begin{equation} \label{6}
    V(t, x) = \rho(t) \Rightarrow \dot{V}(t, x, w) < \dot{\rho}(t), \forall t \in [0, T]
\end{equation}

is sufficient for approximating the reachable set over the course of a trajectory.
By selecting polynomial expressions for $\dot{x}$, $V$, and $\rho$, the constraint in \eqref{6} can be written as:

\begin{equation} \label{7}
\dot{\rho}(t) - \dot{V}(t,x,w) + \lambda_1(t,x,w)[\rho(t) - V(t,x)] + \lambda_2(t,x,w)[t(t-T)] + \sum_{j=1}^{N_w}\lambda_3(t,x,w)g_{w,j}(w)\ \text{is SOS}
\end{equation}
\begin{equation} \label{8}
\lambda_2, \lambda_3\ \text{is SOS}
\end{equation}

where $\lambda_1$, $\lambda_2$, $\lambda_3$ are polynomials with coefficients that are decision variables in the SOS program.
To sustain convexity and handle the bilinear constraints, the algorithm used alternates between two SOS programs: one with decision variables ($\lambda_1, \lambda_2, \lambda_3$) and another with ($V, \rho, \lambda_2, \lambda_3$). This review of the methodology used in \cite{funnel-libraries} is simplified and deliberately leaves out other details such as the decision variables associated with the cost function.

Sum of square optimization can also be used in the performance analysis of black-box algorithms that are widely used in machine learning with large datasets.
The authors in \cite{tan2021analysis} utilize the sum of squares programming to provide convergence rate bounds for first-order optimization algorithms.

\subsection{Multiparametric Programming}
Multiparametric programming is a powerful methodology to compute the solution sets to optimization problems with parametric uncertainty in the right-hand side of the constraints or in the objective function without making any assumptions about the underlying data distributions.
Multiparametric programming can be used to solve optimal solutions to problems under uncertainty and also problems with feasible regions where the solutions to all sub-regions of the feasible regions are desired \cite{multiparametric-programming-process-systems}.
The general multiparametric programming problem is formulated by:

\begin{mini}|s|[0]
    {x \in \mathbb{R}^n, \theta \in \mathbb{R}^m}{f(x,\theta)}
    {\label{9}}{}
    \addConstraint{g(x,\theta)}{\leq 0}
    \addConstraint{h(x, \theta)}{= 0}
\end{mini}

where $x$ is the vector of decision variables and $\theta$ is the vector of uncertain parameters.
Multiparametric programming builds off of the Basic Sensitivity Theorem and uses the result that the active set is constant in the neighborhood of a realization of the uncertain parameter vector $\theta$.
From this, critical regions with constant active sets within the parameter space of the uncertain vector can be constructed.
An explicit solution to each critical region can be derived from the associated unique set of KKT conditions \cite{multiparametric-programming-process-systems}.

\begin{equation}
x^*(\theta) = \left\{ 
  \begin{aligned}
  &x_1(\theta) \; \text{if} \; \theta \in \theta^1  \\ 
  &x_2(\theta) \; \text{if} \; \theta \in \theta^2  \\ 
  &\;\;\;\;\;\;\;\;\;\;\; \vdots \\
  &x_3(\theta) \; \text{if} \; \theta \in \theta^3  \\ 
  \end{aligned}
\right\}
\end{equation}

One of the most famous applications of multiparametric programming is in the development of explicit MPC \cite{explicit-mpc}. The generic MPC problem is to solve the following finite horizon regulation problem with every control tick:

\begin{mini}|s|[0]
    {x \in X, u \in U}{\sum_{k=0}^{N-1}l(x_k, u_k) + F(x_n)}
    {\label{10}}{}
    \addConstraint{x_{k+1}}{= f(x_k, u_k)}
    \addConstraint{x_0}{= x(t)}
    \addConstraint{u_k}{= K(x_k)}
\end{mini}

where $X \subseteq \mathbb{R}^n$ and $U \subseteq \mathbb{R}^m$ are closed sets containing the origin.
$f$ represents the system dynamics, $l$ is the optimal cost-to-go, $F$ is the terminal cost, and $K$ is some state feedback gain \cite{Alessio2009}.

\cite{explicit-mpc} showed that the MPC problem \eqref{10} with quadratic cost and linear time-invariant system dynamics has an explicit solution that is continuous piecewise affine in decision variables, continuous piecewise quadratic in the objective function, and has polytopic critical regions.
This allows for the offline computation of the optimal control law for the entire bounded state space. The online computation becomes simplified and is just a matter of determining what critical region of the state space the current state is in.

\subsection{Stochastic optimization}
Stochastic optimization allows a user to specify the probability distributions that uncertain parameters come from.
Probability distributions can be placed on variables in the constraints, objective, or both.
A generic formulation for a stochastic optimization program that optimizes for the expected values of random variables can be written as:

\begin{mini}|s|[0]
    {x} {\mathop{\mathbb{E}}[f(x, w)]}
    {\label{0}}{}
    \addConstraint{\mathop{\mathbb{E}}[g(x, w)]}{\leq 0}
    \addConstraint{\mathop{\mathbb{E}}[h(x,w)]}{= 0}
\end{mini}

where $w$ represents the uncertain variables.
Depending on the probability distributions that are used in the problem, stochastic optimization can struggle with tractability.
However, for certain types of probability distributions, expectations are relatively cheap and the optimization problem can be efficiently solved.
We now discuss a couple of ways in which stochastic optimization has been effectively utilized for trajectory planning of dynamical systems with uncertain dynamics and/or uncertainty in the environment.
The first method proposed in \cite{gaussian-planning} assumes all uncertainty is Gaussian and linear dynamics:

\begin{equation} \label{14}
    x_{t+1} = Ax_t + Bu_t + w_t + \nu_t
\end{equation}

where $w_t \sim N(0, Q)$ represents model uncertainty and $\nu_t \sim N(0, R)$ represents external disturbances.
Under the assumptions shown in \eqref{14} and that the initial state is a Gaussian distribution $N(x_0, P_0)$, the mean and covariance of the Gaussian distributions of future states can be represented as:

\begin{equation} \label{15}
    \mu_t = \sum_{i=0}^{t-1}A^{t-i-1}Bu_i + A^tx_0
\end{equation}

\begin{equation} \label{16}
    \Sigma_{x_t,y_t} = \sum_{i=0}^{t-1}A^{t-i-1}Q(A^T)^{t-i-1} + \sum_{i=0}^{t-1}A^{t-i-1}R(A^T)^{t-i-1} + A^tP_0(A^T)^t
\end{equation}

From \eqref{15} and \eqref{16} we can see that the future state means is a linear function of control inputs and the future state covariance is independent of control inputs and is known a priori.
This allows the stochastic obstacle-free trajectory planning problem with chance constraints on hitting an obstacle to be written as a deterministic mixed integer linear program.

To handle non-Gaussian uncertainty, \cite{nongaussian-sampling} proposes a sampling scheme for systems with future states that depend explicitly on the initial state $x_0$, control inputs $u$, and additive uncertainty $
\nu$:

\begin{equation} \label{17}
    x_t = \sum_{i=0}^{t-1}A^{t-i-1}B(u_i + \nu_i) + A^tx_0
\end{equation}

The general approach is to sample N pairs where each pair is composed of an initial state $x_0$ and vector of additive uncertainty \{$\nu_0, ..., \nu_{T-1}$\}. Using these N pairs, the intractable stochastic optimization problem is turned into a tractable deterministic one where the percentage of the N pairs that succeed approximates the success rate of the optimized control input trajectory.

To handle non-Gaussian probability distributions without sampling, it has been shown that moments of probability distributions can be utilized to model the uncertainty and formulate convex trajectory optimization programs \cite{risk-bounded-trajectories}.

We refer the reader to the survey papers \cite{optimization-under-uncertainty-soa} and \cite{stochastic-programming-process-systems} for more comprehensive overviews of stochastic optimization.

\subsection{Loop formulations}
Perhaps the simplest model for epistemic uncertainty is the interval model \cite{shariatmadar2021interval}, where one only provides two values to capture the uncertainty of an uncertain parameter: the lower and upper bound. Therefore, optimization problems utilizing intervals on some of the variables is a form of optimization under epistemic uncertainty. One approach to optimizing with interval variables is a nested loop formulation:

\begin{equation}
\label{nested-loop}
\begin{aligned}
& \underset{x}{\text{min}} \; \underset{w}{\text{max}}
& & f(x, w) \\
& \text{subject to}
& & g(x, w) \leq 0, \\
& & & h(x, w) = 0, \\
& & & x \in X, \\
& & & w_{lb} \leq w \leq w_{ub} \\
\end{aligned}
\end{equation}

where the inner loop is a search over the epistemic variables $w$ for the upper bound on the objective and the outer loop is the true optimization problem at hand for chosen values of the epistemic variables \cite{portfolio-optimization-epistemic}.
$x$ is the vector of decision variables, $g(.)$ and $h(.)$ represent generic inequality and equality constraints, respectively.
A nested loop formulation for optimization with interval variables is a form of robust optimization and the similarities with minimax MPC as mentioned earlier in this paper should be apparent.

A nested loop optimization problem is computationally intensive due to the need to solve the inner optimization problem at every step of the outer optimization problem. Decoupled loop methods such as the sequential optimization reliability assessment (SORA) method aim to reduce the computational intensity of nested loop methods \cite{reliability-based-design}.
The decoupled loop formulation of \eqref{nested-loop} can be written as:

\begin{equation}
\label{decoupled-one}
\begin{aligned}
& x^* = \underset{x}{\text{argmin}}
& & f(x, w^*) \\
& \text{subject to}
& & g(x, w^*) \leq 0, \\
& & & h(x, w^*) = 0, \\
& & & x \in X, \\
\end{aligned}
\end{equation}

\begin{equation}
\label{decoupled-two}
\begin{aligned}
& w^* = \underset{x}{\text{argmax}}
& & f(x^*, w) \\
& \text{subject to}
& & g(x^*, w) \leq 0, \\
& & & h(x^*, w) = 0, \\
& & & w_{lb} \leq w \leq w_{ub} \\
\end{aligned}
\end{equation}

In this decouple formulation, the two optimization problems are solved iteratively with the epistemic variables $w$ being fixed in the optimization problem \eqref{decoupled-one} and the decision variables $x$ being fixed in the optimization problem \eqref{decoupled-two} \cite{portfolio-optimization-epistemic}.
In decoupled loop methods, there are still two optimization loops but since the loops are not nested, the comparative computational efficiency can be significant.
Decouple loop methods are popular for robustness and reliability-based design optimization problems \cite{robustness-bases-design} \cite{reliability-based-design}.

Sometimes it is possible to convert a double loop formulation, whether it be nested or decoupled, into a single loop formulation.
\cite{likelihood-representation-of-epistemic-uncertainty} proposed a single loop formulation to robust optimization problems with interval uncertainty by estimating a unique distribution for the random variables via a worst-case maximum likelihood-based estimation.
\cite{portfolio-optimization-epistemic} extended the work in \cite{likelihood-representation-of-epistemic-uncertainty} to incorporate the correlation between input random variables.
The approach is to first obtain a unique distribution for the random variables via a nested optimization problem:

\begin{equation}
\label{single-loop-one}
\begin{aligned}
& \underset{p}{\text{max}} \; \underset{w}{\text{min}}
& & \text{log}(L(p;w)) \\
& \text{subject to}
& & w_{lb} \leq w \leq w_{ub} \\
\end{aligned}
\end{equation}

where $p$ is the parameters of a multivariate normal distribution, $\mu$ and $\Sigma$. $\text{log}(L(p, w))$ is the log-likelihood function for the multivariate normal distribution of the random variable $w$.
After solving the above optimization problem, the resulting PDF that represents the random variables under interval uncertainty can then be used in a single loop optimization formulation, such as \eqref{decoupled-one}, where $w^*$ is chosen to be worst-case maximum likelihood estimates.

\subsection{Bayesian inference}

Although Bayesian methods typically only apply to uncertainty distributions that are completely known, Bayesian inference has been used within optimization methods that incorporate epistemic uncertainty \cite{reliability-based-design-mixed-aleatory-epistemic}.
An important application of epistemic optimization is reliability-based design.
In reliability-based design, the reliability for a particular constraint can be written as:

\begin{equation}
\begin{aligned}
    R = Pr[g(X, P) > 0]
\end{aligned}
\end{equation}

where $X$ and $P$ are vectors that contain both aleatoric and epistemic variables.
Due to the epistemic variables, the reliability $R$ is uncertain and can be modeled using Bayesian inference.
See \cite{reliability-based-design-mixed-aleatory-epistemic} for a review of Bayesian inference as it applies to reliability.

In \cite{bayesian-reliability-optimization} a confidence measure is defined $\zeta(\mu_x)$ and proposes a multi-objective optimization problem for reliability-based optimization under epistemic uncertainty:

\begin{equation}
\begin{aligned}
& \underset{\mu_x}{\text{min}} \; f(\mu_x, \mu_p)\; \underset{\mu_x}{\text{max}} \; \zeta(\mu_x) \\
& \text{subject to} \; 0 \leq \zeta(\mu_x) \leq 1
\end{aligned}
\end{equation}

Solving this problem results in a set of designs with different confidence values for the desired reliability value $R$.
To make the above problem more tractable, one usually selects a confidence value and then the reliability is computed from the reliability distribution produced via Bayesian inference \cite{reliability-based-design-mixed-aleatory-epistemic}.
This leads to the Bayesian reliability-based design optimization formulation:

\begin{equation}
\begin{aligned}
& \underset{d, \mu_x}{\text{min}}
& & f(d, \mu_x, \mu_p) \\
& \text{subject to}
& & Pr(g(d, X, P) \leq 0) \leq Pr_{target} \\
& & & h(d) \geq 0 \\
& & & d_L \leq d \leq d_u, \; \mu_{xL} \leq \mu_x \leq \mu_{xU} \\
\end{aligned}
\end{equation}

where $d$ is a vector of deterministic variables, $X$ is a vector of aleatoric variables, $P$ is a vector or epistemic variables, and $g(d, X, P) \leq 0$ defines a failure region \cite{reliability-based-design-mixed-aleatory-epistemic}.

\subsection{Upper and lower expectations}

When optimizing with variables that are modeled by epistemic models such as the p-boxes, a typical approach is to take a lower or upper expectation of those epistemic variables and then optimize the resulting deterministic problem.
Whether to use the lower or upper expectation is decided by if the objective is to minimize or maximize and whether one wants to optimize for the worst or best-case scenario.
A lower expectation can be formulated as:

\begin{equation}
\begin{aligned}
E_l = \underset{p \in P}{\text{min}} \;
& \int_{W} (h(w) \in H) p(w) \; dw\\
\end{aligned}
\end{equation}

where $h(w)$ is the value of interest that depends on the uncertain variable $w$ from the space of uncertain variables $W$.
$p$ is a probability distribution within the set of distributions $P$.
\cite{epistemic-trajectory} presents a method to generate optimal trajectories that are robust against epistemic uncertainty where the uncertainty is modeled with p-boxes.
They utilize a surrogate model and take the lower expectation of the objective and constraints to solve the epistemic optimization problem.
We refer the reader to \cite{linear-programming-p-box} for a theoretical treatment of the solution to linear programs with p-box uncertainty models.
Here, the approach is to convert the uncertain optimization problem to a deterministic one using imprecise decision theory.

\subsection{Transform to deterministic problem}
The general approach of taking an uncertain optimization problem and transforming it into a deterministic one accounts for many of the methods in epistemic optimization.
The key distinguishing feature of these different approaches to epistemic optimization lies in the method of selecting fixed variables/functions in place of epistemic ones.
Here we discuss a few of these approaches that are apparent in recent literature.
Perhaps the most simple is to approximate the epistemic parameters through some statistical function \cite{dealing-with-epistemic-multiobjective} \cite{objective-penalty-function}:

\begin{equation}
    g(f(x, w))
\end{equation}

where $g(.)$ is some statistical function, $f(.)$ is the function to approximate, and $w$ is an epistemic variable.
The simplicity of this approach may come at the cost of producing inaccurate solutions.
For generic optimization problems, it is not surprising that reducing an epistemic parameter down to a fixed parameter via a statistical function may lead to an optimization problem that uses many inaccurate parameters.
Another approach is to define a robustness criterion $R$ and maximize this criterion.
$R$ is defined by the variation of the associated uncertain function $f(x)$ which has uncertainty in $x$ \cite{robustness-multi-objective}.
Defining a robustness criterion is popular in the field of robust design and we refer the reader to the comprehensive survey paper \cite{robust-design-survey} to learn more about optimization under uncertainty as it applies to design.
Another category is interval-based approaches where the uncertainty is captured by an upper and lower bound:

\begin{equation}
    f(x, w) = [y_l, y_u]
\end{equation}

\cite{multiobjective-fuzzy-system} presents a genetic algorithm for optimization problems where uncertainty in the objective is represented by intervals of fuzzy sets.
See \cite{dealing-with-epistemic-multiobjective} for a survey on epistemic multi-objective optimization.

\subsection{Fuzzy optimization}

\subsubsection{Main Approaches to Fuzzy Optimization}
The fuzzy optimization problems considered as $A=\{a\}$ is a set of possible outcomes, and an objective function is defined as $f:A\longrightarrow\mu(B)$, where $\mu(B)$ is the fuzzy sets defined in $B$, the real line. In other words, $f(a)$ is a fuzzy value which illustrates a fuzzy evaluation of the possible outcome $a\in A$. The set of possible options is defined by a fuzzy set $F$ in $A$ such that $F(a)\in[0,1]$ is called the degree of possibility. For fully possible the degree is 1 and 0 otherwise, through all values.

\subsubsection{Mathematical form of the fuzzy optimization problem} 
The fuzzy optimization problem is defined as
\begin{equation}\label{fuzzy-eq}
\begin{aligned}
& \underset{x\widetilde{\in} F}{\widetilde{\max}}
& & f(x)\\
\end{aligned}
\end{equation}

where it is about finding a possibly maximum value of $f$ ($\widetilde{\max}$) over the $x$'s “possibly belonging” ($\widetilde{\in}$) to the fuzzy feasible set $F$. The problem \eqref{fuzzy-eq} are discussed in various formats, e.g., given by Bellman and Zadeh’s \cite{fuzzy-environment}.

A major drawback of fuzzy optimization is that implementation is not efficient. Currently one must use fuzzy arithmetic or fuzzy operators which transforms the problem to be solved under fuzzy logic \cite{fuzzy-arithmetic-industrial} \cite{lp-interval-fuzzy-sets}. This defuzzification process is not efficient enough yet for online use and is a possible research direction.

\subsection{Constrained optimization under uncertainty using decision theory} 
One of the most general models of epistemic uncertainty, even more, general than interval and belief function models, is imprecise probability. Generally, an optimization problem under (imprecise) uncertainty is defined as follows
\begin{align}\label{COUU}
&\underset{x{\in} \mathcal{X}}{{\max}}\quad f(x,Y)\notag\\
&\text{s.t. }\quad xRZ,
\end{align}
where $x$ is the optimization vector in any set $\mathcal{X}$, $Y$ and $Z$ are random vectors in $\mathcal{Y}$ and $\mathcal{Z}$, and $R$ is a relation in the set $\mathcal{X}\times\mathcal{Z}$. A simple case of constrained optimization under an uncertainty problem is a linear programming problem under uncertainty which is defined as
\begin{align}\label{LPUU}
&\underset{x\in\mathbb{R}^n_{\ge 0}}{{\max}}\quad U^Tx\notag\\
&\text{s.t. }\quad Yx\ge Z,
\end{align}
where $x$ is the optimization vector in $\mathbb{R}^n_{\ge 0}$, $(Y,Z,U)$ are random vectors in $\mathbb{R}^{m\times n}\times\mathbb{R}^m\times\mathbb{R}^n$, and $R:=\ge$ is the relation in the set $\mathbb{R}^m\times\mathbb{R}^m$. Many kinds of research and works, like \cite{shariatmadar2019pbox, shariatmadar_contamination_2020,shariatmadar2021interval} have been done to solve the optimization under uncertainty problems \eqref{COUU} and \eqref{LPUU} at our group in KU Leuven. We first convert the problem to a decision problem and use decision criteria to provide solutions based on the criteria.

\section{Discussion}
\subsection{Comparison}
In this review, a variety of different approaches to optimization under aleatoric and epistemic uncertainty were discussed.
When the distributions of the random variables are precisely known, the uncertainty is classified as aleatoric.
Aleatoric uncertainty is inherent to the process under consideration.
If an optimization problem only considers aleatoric uncertainty then the optimization problem can typically be categorized as a stochastic optimization program.
Stochastic optimization leverages knowledge of the underlying probability distributions of the random variables in the problem.
Once probability distributions for the random variables are obtained, one can use those distributions to take the expectations of the variables and optimize for the average scenario.
If sampling from the distributions is cheap, one can also optimize over a set of samples.
Furthermore, the uncertainty moments of the random variables can be utilized to introduce chance constraints or constraints that must be satisfied with a given level of confidence \cite{chance-constrained-programming}.

When the distributions of the random variables are not precisely known, the uncertainty is classified as epistemic.
Epistemic uncertainty comes from a lack of knowledge and can, in principle, be reduced to aleatoric uncertainty after sufficient experimentation.
The type of epistemic uncertainty model used in an optimization problem influences the procedure to solve the problem.
A simple and popular method for modeling epistemic uncertainty is through a bounded set where the random variable is known to lie within the set but no probabilities are associated with the set.
Robust optimization uses bounded sets to optimize for the worst-case scenario, leading to conservative behavior.
Multiparametric programming also uses bounded sets and computes offline the optimal solutions as a function of all possible realizations of uncertain variables.
Multiparametric programming suffers from the curse of dimensionality and has thus limited multiparametric programming to relatively simple systems operating in more controlled environments.
The interval model is a bounded set in one dimension and when using interval models, one can utilize double or single-loop formulations.
When using the more informative p-box model, it is most common to take the lower/upper expectations of the random variables and then optimize using those expectations.
Many epistemic optimization methods optimize for the worst-case realization of the epistemic variables but one can also use Bayesian inference methods to enforce levels of reliability/robustness into the problem.
A separate class of optimization problems, fuzzy optimization, exists for optimization problems using fuzzy sets.
The main drawback of fuzzy optimization is that the current solution methods are not efficient enough for real-time use.

\subsection{(Constrained) optimization under uncertainty in machine learning}
Generally, when there is data available or a repeatable task is running one question is that could we learn a pattern from the data or the tasks. Mainly, in machine learning, we use this data as input to learn or train a model. This has been done via several machine learning techniques such as supervised or unsupervised learning, reinforcement or inverse reinforcement learning, and so on. In almost all of these techniques, we use optimization theory to solve machine learning problems. For instance, in regression problems where the goal is to fit a model to data, one solution is to use a Bayesian neural network \cite{kononenko1989bayesian}, in which the parameters are represented by probability distributions. In this method, a back-propagation method is used to minimize a loss function. In one of our ongoing research, we solve an interval neural network \cite{ishibuchi1993, oala2021} problem via a constrained optimization problem under uncertainty. By doing so, we find the best model with the highest accuracy to fit the data using the worst-case scenario optimization technique. This work is in progress and will be our next publication.

\section{Conclusion}

The field of optimization has made tremendous strides in progress over the last half-century.
Theoretical developments have improved our understanding of the structure of optimization problems.
Advancements in computer technology continue to make optimization more applicable to everyday engineering.
With the inclusion of uncertainty into the optimization problems, the resulting solutions are more robust to the inevitable discrepancies between the modeled parameters and the real system.
However, simply assigning probability distributions to unknown parameters often assumes more knowledge than what one really has.
By accepting the lack of knowledge about certain parameters, the problem then becomes one of epistemic uncertainty.
Using epistemic models often results in more realistic models, however, it complicates the problem significantly.
Incorporating epistemic uncertainty can immediately make a tractable optimization problem intractable.
The most popular model for epistemic uncertainty is the bounded set.
It is apparent that the theory of optimization under epistemic uncertainty with more advanced models than the bounded set needs to be further developed.
There is a disproportionate amount of literature around advanced epistemic uncertainty modeling and the associated optimization problems.
A unified mathematical framework for optimizing under epistemic uncertainty beyond the bounded set is of paramount importance and is a research direction with many opportunities.
Furthermore, the need to develop efficient algorithms to solve optimization under epistemic uncertainty will naturally follow once the theory is in place.

\section{Acknowledgment}
This work is supported by the FETOPEN European Union's Horizon 2020 research and innovation programme under grant agreement No. 964505 (\href{https://www.epistemic-ai.eu/}{Epistemic AI}).

\medskip

\printglossary
\printbibliography
\end{document}